\pdfoutput=1

\documentclass[11pt]{article}

\usepackage{acl}

\usepackage{times}
\usepackage{latexsym}

\usepackage[T1]{fontenc}

\usepackage[utf8]{inputenc}

\usepackage{microtype}

\usepackage{inconsolata}

\usepackage{graphicx}

\usepackage{xspace}
\usepackage{soul}
\newcommand{\name}{\textit{ContextLens}\xspace}
\usepackage{tcolorbox}
\definecolor{cYellow}{RGB}{255,255,3}
\definecolor{cBlue}{RGB}{69,123,157}
\definecolor{cRed}{RGB}{231,56,71}
\definecolor{cRed_1}{RGB}{191,30,46}
\definecolor{cGray}{RGB}{168,218,219}
\definecolor{cBlue_2}{RGB}{5,48,97}
\definecolor{cBlue_1}{RGB}{115,186,214}
\definecolor{cBlue_3}{RGB}{13,76,109}
\definecolor{cBlue_4}{RGB}{64,121,160}
\definecolor{cOrange}{RGB}{250,134,0}
\definecolor{cBlue_6}{RGB}{13,76,109}
\definecolor{cBlue_7}{RGB}{16,106,130}
\definecolor{cBlue_8}{RGB}{19,136,160}
\definecolor{cBlue_9}{RGB}{115,184,214}
\definecolor{stepcolor}{HTML}{d79b00}
\definecolor{contentcolor}{HTML}{3439a2}

\usepackage{booktabs}
\usepackage{multirow}
\usepackage{amsmath,amssymb}
\usepackage[table]{xcolor}
\usepackage{colortbl}
\definecolor{tablegray}{gray}{0.5}
\title{ContextLens: Modeling Imperfect Privacy and Safety Context for Legal Compliance}

\author {
    {\bf Haoran Li}\textsuperscript{\rm 1,2},
    {\bf Yulin Chen}\textsuperscript{\rm 3},
    {\bf Huihao Jing}\textsuperscript{\rm 2},
    {\bf Wenbin Hu}\textsuperscript{\rm 2},
    {\bf  Tsz Ho Li}\textsuperscript{\rm 2},
    \\
    {\bf  Chanhou Lou}\textsuperscript{\rm 4},
    {\bf  Hong Ting Tsang}\textsuperscript{\rm 2},
    {\bf  Sirui Han}\textsuperscript{\rm 2}\thanks{ Corresponding Author},
    {\bf Yangqiu Song}\textsuperscript{\rm 2}\\
    \textsuperscript{\rm 1}Beihang University, 
    \textsuperscript{\rm 2}HKUST, 
    \textsuperscript{\rm 3}National University of Singapore, \\
    \textsuperscript{\rm 4}Faculty of Law, University of Macau\\
    \texttt{\{hlibt, hjingaa, whuak, thliai\}@connect.ust.hk}, \texttt{chenyulin28@u.nus.edu}\\ 
    \texttt{siruihan@ust.hk}, 
    \texttt{chlou@um.edu.mo},
    \texttt{yqsong@cse.ust.hk}\\
}

\begin{document}
\maketitle
\begin{abstract}
Individuals' concerns about data privacy and AI safety are highly contextualized and extend beyond sensitive patterns.
Addressing these issues requires reasoning about the context to identify and mitigate potential risks.
Though researchers have widely explored using large language models (LLMs) as evaluators for contextualized safety and privacy assessments, these efforts typically assume the availability of complete and clear context, whereas real-world contexts tend to be ambiguous and incomplete.
In this paper, we propose \name, a semi-rule-based framework that leverages LLMs to ground the input context in the legal domain and explicitly identify both known and unknown factors for legal compliance.
Instead of directly assessing safety outcomes, our \name instructs LLMs to answer a set of crafted questions that span over applicability, general principles and detailed provisions to assess compliance with pre-defined priorities and rules.
We conduct extensive experiments on existing compliance benchmarks that cover the General Data Protection Regulation (GDPR) and the EU AI Act.
The results suggest that our \name can significantly improve LLMs' compliance assessment and surpass existing baselines without any training.
Additionally, our \name can further identify the ambiguous and missing factors.

\end{abstract}

\section{Introduction}
\label{sec: intro}

The integration of tool use and agentic workflows empowers LLMs with unprecedented accessibility to handle a wide range of applications~\cite{MCP, berkeley-function-calling-leaderboard}.
Such unrestricted access control and unpredictable responses worsen people's safety and privacy concerns.
On one hand, models' outputs may cover sensitive or harmful information that should not be disclosed.
On the other hand, automatic agentic workflows may amplify the malicious intentions without notice.
Existing practices commonly apply filtering and safety alignment techniques to protect privacy and safety. 
Filtering identifies sensitive patterns via regular expressions and simple classifiers.
Safety alignment curates tailored data to harness LLMs to safety, value, and privacy requirements~\cite{Christiano-2017-rlhf, rafailov2023direct, inan-2023-llama-guard}.
However, both methods fail to cover people's full concerns and generalize well to evolving attacks~\cite{chen2025can, LI-2023-Jailbreak}.

Motivated by the fact that people's perceptions of safety and privacy are highly contextualized~\cite{Nissenbaum-2010-CI} and LLMs can serve as good judges, recent works~\cite{shvartzshnaider2024llm, ghalebikesabi-2024-operationalizing, cheng-2024-cibench, fan2024goldcoin} start to leverage LLM-as-a-judge to identify potential safety and privacy issues inside a given context.
Various prompting strategies, including retrieval augmented generation (RAG) and task-specific model fine-tuning strategies~\cite{hu2025context, li-2024-privacychecklist}, are explored to enhance LLMs' contextual judgment capabilities.
In addition, several benchmarks~\cite{ kang2025polyguard, li2025privaci, cheng-2024-cibench, mireshghallah2024can} collect real and synthetic data with references to existing safety policies, regulations and standards to evaluate LLMs' contextual safety and privacy judgment abilities.

Though these strategies' effectiveness has been validated in several benchmarks, they always assume that the given context is perfect and sufficient for the LLM to judge violations of existing safety and privacy standards.  
On the contrary, in practice, a short textual description of context tends to be incomplete and ambiguous~\cite{yi2025privacy}.
Suppose the context describes an individual who gives its consent to ChatGPT's ``Terms and Conditions'' to use ChatGPT.
Most LLM judges will deem this consent compliant with existing standards.
However, if the individual is a child under 13 residing in the European Union, such consent does not constitute a lawful basis according to the General Data Protection Regulation (GDPR).
In this case, the context is lawful only if the child's legal guardians provide authorized consent.
Otherwise, the child may be at risk of being exposed to inappropriate AI-generated content.
This example partly explains why ChatGPT faced a temporary ban in Italy for non-compliance with the GDPR\footnote{https://www.reuters.com/technology/italy-fines-openai-15-million-euros-over-privacy-rules-breach-2024-12-20/}.

To address the context incompleteness and ambiguity, we propose \name, a semi-rule-based method that explicitly models the unknown and vague factors inside a given textual description.
Our \name decomposes the regulation into distinct components, including scope, lawfulness, general principle and special conditions.
Instead of leveraging LLM-as-a-judge to generate compliance outcomes, we prompt LLMs with each separate component to generate structured outputs.
These structured outputs allow us to identify whether the covered sub-rules are absent in the context.
By aggregating responses across components, we apply predefined rules to assess privacy and safety compliance and highlight the ambiguous and unknown contextual factors that may potentially affect the judgments.
We conduct extensive experiments to demonstrate the effectiveness of \name with the new state-of-the-art compliance performance.
In summary, our contributions are as follows:\footnote{Code and data are publicly available at \url{https://github.com/HKUST-KnowComp/ContextLens}.}


1) We identify the context incompleteness and ambiguity for compliance assessment, a practical yet under-estimated weakness.

2) We present \name, the first semi-rule-based approach to assess privacy and safety compliance following the hierarchical nature of established legal statutes on safety and privacy.

3) Beyond deriving the outcome, \name is the first framework to explicitly identify missing and ambiguous contextual factors that may further affect the compliance assessment.

4) Extensive evaluations are conducted to show that \name outperforms all existing baselines and identify that nearly 100\% of current benchmark data suffer from imperfect context.

\section{Related Works}
\label{sec: relate}

\textbf{Common Practices to Evaluate Privacy and Safety}.
Existing literature frequently adopts regular expressions to match the exact sensitive keywords to calculate the attack success rate (ASR)~\cite{zou2023universal, chen2024defense}. 
The ASR is widely used to measure the attack effectiveness, including but not limited to jailbreak, prompt injection and training data extraction.
In addition, recent safety classifiers, such as OpenAI Moderation API~\cite{markov2023holistic}, ShieldGemma~\cite{zeng2024shieldgemma}, Llama Guard~\cite{inan-2023-llama-guard}, Guardreasoner~\cite{liuyue_GuardReasoner} and Prompt-Guard~\cite{meta2024-prompt_guard}, are specially post-trained on safety datasets to identify potential risks given input prompts or model responses.
However, these evaluation methods are typically task-specific and fail to generalize to broader safety concerns or align with existing regulatory standards.

\textbf{Contextualized Privacy and Safety Evaluation}.
Recently, a few works have started to assess contextualized privacy and safety with the Contextual Integrity (CI) theory~\cite{Nissenbaum-2010-CI}.
CI states that five contextualized parameters including sender, recipient, information subject, information types (transmitted attributes, topics and other sensitive information about the subject), and transmission principle, can shape the privacy context during information transmission~\cite{Benthall-CI-2017}. 
\citet{Barth-2006-CI} transform the context into formal logic languages to explicitly model the context.
Another line of work~\cite{shvartzshnaider2024llm, shvartzshnaider2025position, ghalebikesabi-2024-operationalizing, cheng-2024-cibench, li2025privaci, fan2024goldcoin} leverages LLMs to annotate the context into structured CI templates and reuses the LLM-as-a-judge to determine evaluation results.
All of these works assume that a perfect context is provided.
~\citet{yi2025privacy} study the context ambiguity inside existing privacy reasoning benchmarks~\cite{mireshghallah2024can, shao2024privacylens} and propose Camber to augment extra synthetic context for disambiguation.
In contrast, our \name adopts the rule-based method to assess privacy and safety compliance under real-world assumptions where introducing synthetic context is not applicable.



\section{ContextLens Construction}
\label{sec: data}

\begin{figure*}[t]
\centering
\includegraphics[width=0.997\textwidth]{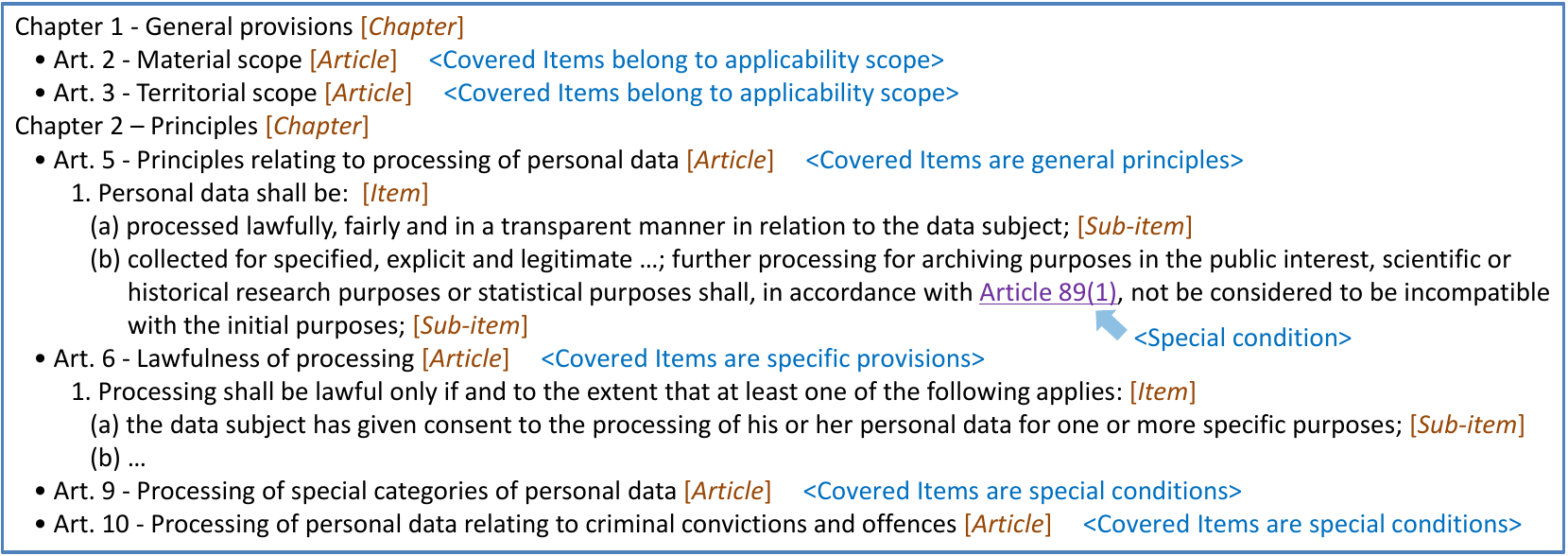}
\vspace{-0.3in}
\caption{
An example of \name' regulation chunking on the GDPR. 
We decompose the regulation following the chapter$\rightarrow$article$\rightarrow$item$\rightarrow$sub-item structure.
We leverage LLM-as-a-judge to perform legal document chunking at the item and sub-item level with human annotator verification.
}
\label{fig:example}
\vspace{-0.23in}
\end{figure*}

\subsection{Motivations}
\label{sec: motivation}

Our \name is motivated by two factors, including both recent experimental findings and the hierarchical nature of current legal statutes.

Recent experimental findings suggest that most LLMs tend to hallucinate on non-existent situations for contextualized privacy evaluations.
This tendency is further exacerbated when the provided context is incomplete~\cite{yi2025privacy,li2025privaci}.
What is worse, in retrieval-augmented generation (RAG), existing retriever implementations in the legal domain suffer from low recall with Recall@5 rates below 10\%~\cite{hou-etal-2025-clerc}.
Consequently, irrelevant or non-applicable retrieved rules can mislead LLMs into hallucinating on non-existent context to attempt to fit those rules.

Moreover, established legal statutes are systematically organized with hierarchical rules to resolve potential conflicts.
These conflicts become acute when broad statutory principles overlap with narrowly defined exceptions, or when newly enacted provisions interact with longstanding rules~\cite{Michaels-2020-Conflict, kornhauser1994adjudication}. 
Judicial reasoning in these cases relies on a hierarchy of legal maxims, including lex superior derogat legi inferiori (higher law prevails over lower), lex specialis derogat legi generali (the specific prevails over the general), and lex posterior derogat legi priori (the later prevails over the earlier).
For safety and privacy regulations, the hierarchy typically adopts the following order of precedence: \textit{applicability scope} $>$ \textit{special conditions} $>$ \textit{specific provisions} $>$ \textit{general principles}, where ``A $>$ B'' indicates A takes priority over B.
Within a legal regulation, \textit{applicability scope} indicates whether the given context fits into the corresponding material scope and territorial scope.
If the context is not applicable, subsequent chunks do not necessarily apply. 
\textit{Special conditions} are explicitly specified special situations, including exceptional references and provisions related to a specific scenario (i.e., employment, healthcare and minors).
These conditions offer context-dependent provisions that override \textit{common provisions}.
\textit{Common provisions} refers to detailed specifications, including but not limited to permitted actions.
\textit{General principles} represent overarching values, incentives or guidelines used to resolve conflicts.

Inspired by the above findings, our \name divides the regulations into different chunks by assigning each atomic sub-rule to the corresponding priority level.
Instead of directly assessing compliance outcomes, we ask the judge LLM whether sub-rules of each chunk apply to the given context.
If applicable, we let the judge LLM decide whether the context is permitted or prohibited by the given sub-rules.
Lastly, we aggregate all chunks' judgments to determine the context's compliance.
When chunks reach an agreement about their compliance assessment, we directly use it as the final outcome. 
For potential conflicts among chunks, we utilize the hierarchical priority to determine compliance via judicial reasoning.

\subsection{Problem Formulation}
Let $x$ denote the textual description of the context that requires the compliance assessment regarding safety and privacy.
For a given regulation $R$, our \name aims to determine $x$'s compliance outcome $y$ where $y \in \{\text{Permitted, Prohibited, Not Applicable}\}$.
Unlike previous approaches, we assume that $x$ is fixed and that no additional contextual information can be obtained interactively.
Beyond determining $y$, we also seek to identify a set of provisions with missing contextual factors $C = \{c_1, c_2, \dots, c_n\}$ that could potentially influence or even overturn the current compliance assessment $y$.
These contextual factors $c$ generally correspond to detailed legal provisions that involve rich and specific contextual information such as specialized domain information, social roles, and even mental states.
We do not consider conflicts between different regulations.

\subsection{Legal Document Processing}

As shown in Figure~\ref{fig:example}, we first use the identifiers to decompose the legal regulation into a hierarchical structure consisting of \textit{chapter} $\rightarrow$ \textit{article} $\rightarrow$ \textit{item} $\rightarrow$ \textit{sub-item}.
Then, following the aforementioned precedence order that \textit{applicability scope} $>$ \textit{special conditions} $>$ \textit{common provisions} $>$ \textit{general principles}, we perform the document chunking on the EU AI Act and the GDPR at the item and sub-item level.
Figure~\ref{fig:example} provides an illustrative example to show how we chunk the GDPR document.

For \textit{applicability scope} and \textit{general principles}, both are typically outlined at the beginning of a legal document (i.e., Art. 2,3 and 5 in Figure~\ref{fig:example}).
Since the regulations are well structured with the article and item identifiers, we consider the \textit{applicability scope} as disjunctive sub-items such that the input context is deemed applicable when any sub-item applies and the exemptions are not met.
In other words, if the input context satisfies at least one of the sub-items, this context fits into the scope of the corresponding regulation.
On the other hand, \textit{general principles} articulate the fundamental values that should all be respected, so we format these principles as conjunctive terms.
In case the given context is ambiguous to determine compliance for all detailed provisions, then the compliance outcome is carefully assessed by all items within the \textit{general principles}.

For \textit{common provisions} and \textit{special conditions}, we consider items related to obligations, requirements, lawfulness specifications and rights of subjects as \textit{common provisions}, which regulate general behaviors.
In contrast, \textit{special conditions} refer to provisions that address specific situations or exceptional references in \textit{common provisions} and \textit{general principles}.
We use GPT-4o to review the main body of each document to examine all relevant items and sub-items.
Then, three of the authors manually inspect the identified \textit{common provisions} and \textit{special conditions}.
For instance, as shown in Figure~\ref{fig:example}, items under Articles 9 and 10 are about specific situations while  Article 89(1) is an exceptional reference of Article 5(1)(b).
These \textit{special conditions} are regulation-dependent and have higher priority when conflicts with the main body of \textit{common provisions}.
In the GDPR, data processing related to sensitive data, cross-border data transmission and criminal offences are typically governed by specific articles.
On the other hand, the EU AI Act introduces a range of sub-rules to address prohibited AI practices, high-risk AI systems, and general-purpose AI models.

\subsection{Prompt Template Construction to Identify the Imperfect Context}

Prior rule-based methods commonly leverage logical expressions to model the legal document and the input context.
However, when it comes to context ambiguity and incompleteness, the binary true or false assignment may not apply.
In this section, we instruct foundation LLMs to answer with non-committal responses such as ``not sure'' and ``none of the above'' under the structured output constraint to explicitly model the uncertainty inside context.


For the GDPR, we consider items under Articles 8, 9, 10, 11, 44, 86, 87, 88 and 89 as \textit{special conditions} and we decompose \textit{common provisions} into multiple prompt templates covering lawfulness, processor obligations, and rights of the data subject for topic consistency.
For all chunks, each prompt template consists of 4 modules, including the instruction, context, corresponding GDPR article content and output format. 
This output format strictly enforces the LLM to output in JSON format.
A representative prompt is shown as follows:

\begin{tcolorbox}[colback = cBlue_1!10, colframe = cBlue_7,  coltitle=white,fonttitle=\bfseries\small, center title,fontupper=\small,fontlower=\small]
\textbf{Instruction:}
\texttt{<Instruction>}\\
If the article is relevant but does not specify in the context, please answer with "not sure". \\
\textbf{Context:} \texttt{<Context>}\\
\textbf{GDPR Articles:} \\
- Article \texttt{<$d_1$>}: \texttt{<Content\_$d_1$>}\\
- Article \texttt{<$d_2$>}: \texttt{<Content\_$d_2$>}\\
...\\
\textbf{Output Format:} \\
Output format should be in JSON format: \\
\{ \\
\hspace*{1em} ``Article \texttt{<$d_1$>}": ``yes" or ``no" or ``not sure", \\
\hspace*{1em} ``Article \texttt{<$d_2$>}": ``yes" or ``no" or ``not sure", \\
\hspace*{1em} ...\\
\}
\end{tcolorbox}
\vspace{-0.05in}

For the EU AI Act, we follow the official EU AI Act Compliance Checker\footnote{https://artificialintelligenceact.eu/assessment/eu-ai-act-compliance-checker/} to design our prompts.
The compliance checker covers a set of multiple-choice questions related to entity type, scope, and AI system categories, including high-risk and excluded systems.
After answering all the questions, the compliance results are shown with the corresponding provisions. 
These questions naturally align with our identified four-level precedence.
Additionally, we incorporate the ``None of the above'' option inside multiple-choice questions to model the ambiguous or incomplete context.
We first append a prompt template to analyze the context and identify if any AI system is involved to fit into the \textit{applicability scope}.
Then, we concatenate the instruction, original context, analyzed context, list of questions and output format as follows:

\begin{tcolorbox}[colback = cBlue_1!10, colframe = cBlue_6,  coltitle=white,fonttitle=\bfseries\small, center title,fontupper=\small,fontlower=\small]
\textbf{Context Analysis:}
\texttt{<Analysis Instruction>}\\
\textbf{Context:} \texttt{<Context>}\\
\textbf{Output Format for the Analyzed Context:} \\
Output format should be in JSON format: \\
\{ \\
\hspace*{1em}        ``AI\_system\_involved": True/False, \\
\hspace*{1em}        ``AI\_system\_name": ``name of the AI system", \\
\hspace*{1em}           ...\\
\}

\tcblower
\textbf{Instruction:}
\texttt{<Instruction>}\\
\textbf{Context:} \texttt{<Context>}\\
\textbf{Analyzed Context:}
\texttt{<Analyzed Context>}\\
\textbf{List of questions:} \\
question\_1: \texttt{<Content\_$q_1$>}\\
question\_2: \texttt{<Content\_$q_2$>}\\
...\\
\textbf{Output Format:} \\
Output format should be in JSON format: \\
\{ \\
\hspace*{1em} ``question\_1": [option\_1, option\_2, ...], \\
\hspace*{1em} ``question\_2": [option\_1, option\_2, ...], \\
\hspace*{1em} ...\\
\}
\end{tcolorbox}
\vspace{-0.05in}

As illustrated in the output format requirement, our prompt template converts multiple-choice questions into multiple-select questions to accommodate cases where more than one option may apply.
This approach introduces better flexibility and helps reduce token usage with improved inference efficiency.

\subsection{Rule-based Compliance Evaluation}

Our prompt templates are designed to guide LLMs to operate at the item level, without assessing the overall compliance.
To evaluate overall compliance, we employ additional rule-based methods tailored to each regulation.

For the GDPR, we first assess the compliance within each chunk.
For example, in \textit{common provisions} and \textit{general principles}, compliance is granted only when there is no violated sub-rule.
To assess the overall compliance, we use the high-order precedence to overwrite the low-order result.

For the EU AI Act, different options in the multiple-choice questions may lead to the same subsequent questions and compliance result.
To accurately reflect this structure, we re-implement the EU AI Act Compliance Checker as a multi-graph.
Internal nodes represent questions, while edges represent user-selected options.
All the leaf nodes refer to the compliance outcomes accompanied by quoted legal articles.
By traversing the multi-graph from the root question with parsed options, we derive the overall compliance result.

\subsection{Imperfect Context Identification}
In addition to the overall compliance assessment, within each chunk, our rules also track provisions marked as “not sure” or “none of the above.” 
These provisions are viewed as missing contextual factors $c$ of the context.
The presence of such factors indicates that the given context may be imperfect, as they may potentially undermine existing rule-based assessment.
Our subsequent evaluations on these missing factors $c$ suggest that nearly all evaluation samples suffer from imperfect context.

\section{Evaluation Setups}
\label{sec: eval}

\begin{table}[t]
\small
\centering
\resizebox{\columnwidth}{!}{
        \begin{tabular}{l|c|c|c|c}
            \toprule
            \textbf{Domain} & \textbf{Permit} & \textbf{Prohibit} & \textbf{Not Applicable} & \textbf{Total} \\
            \midrule
            GDPR & 150 & 478 & 0 & 628 \\
            AI Act & 202 & 187 & 211 & 600 \\
            
            \bottomrule
        \end{tabular}
        }
        \vspace{-0.1in}
        \caption{Testing data statistics for legal compliance.}
        \label{tab:data_stats}
        \vspace{-0.15in}
\end{table}

\subsection{Evaluated Datasets}
To assess our proposed \name for the legal compliance task, we utilize legal cases from PrivaCI-Bench~\cite{li2025privaci}. 
PrivaCI-Bench collects 6,348 real and synthetic cases, each with compliance outcome annotations, including \textit{Permitted}, \textit{Prohibited}, and \textit{Not Applicable}.
These cases encompass various issues from AI system misuse to data privacy breaches. 
Following PrivaCI-Bench's official data split, we use its testing subsets of the GDPR and AI Act domains. 
The statistics of our evaluated subsets are shown in Table~\ref{tab:data_stats}.

\subsection{Evaluated LLMs and Baselines}
We evaluate the compliance performance on a wide range of closed-source and open-source LLMs, including instruction-tuned LLMs, Large Reasoning Models (LRMs) and LLMs with task-specific fine-tuning.
For instruction-tuned LLMs, we evaluate Qwen2.5-7B-Instruct~\cite{Yang2024Qwen2TR}, GPT-4o-mini, and GPT-4o.
For LRMs, given the same prompt as instruction-tuned LLMs, LRMs typically employ long Chain-of-Thought reasoning strategies, which generate the intermediate reasoning procedures before producing the final answers.
We evaluate DeepSeek-R1 (671B)~\cite{guo2025deepseek}, Gemini-2.5-Flash~\cite{geminicard}, and o3-mini~\cite{o3mini}.
For task-specific fine-tuning, we report the performance of ContextReasoner-7B~\cite{hu-2025-contextreasoner}, which is trained on the OpenThinker-7B~\cite{openthoughts2025}.
OpenThinker-7B is a supervised fine-tuned (SFT) LLM based on Qwen2.5-7B-Instruct, trained on OpenThought-114k~\cite{openthoughts2025} dataset.
ContextReasoner-7B models further fine-tune OpenThinker-7B using PrivaCI-Bench's training data through supervised fine-tuning (ContextReasoner-7B-SFT) and the proximal policy optimization (ContextReasoner-7B-PPO) algorithm~\cite{schulman2017ppo} to achieve optimal compliance performance.

To compare our \name with these baselines, we reuse the exact multiple-choice prompting template used by ContextReasoner, which is shown in the Appendix.
Our evaluated setups can be categorized as follows:

\textbullet  \textbf{Direct}: We use the multiple-choice template to evaluate the instruction-tuned LLMs' compliance performance. 

\textbullet  \textbf{Long CoT}: We use the same multiple-choice template to evaluate the LRMs' performance. 

\textbullet  \textbf{Fine-tuned}: We apply the same multiple-choice template to evaluate the compliance performance on ContextReasoner models after training on PrivaCI-Bench's training subsets.

\textbullet  \textbf{RAG}: Given the input case context, we leverage the LLMs' legal knowledge to interpret the scenario and retrieve relevant legal rules via BM25, incorporating both into the prompt for retrieval-augmented compliance reasoning, consistent with the RAG implementation in PrivaCI-Bench.

\textbullet  \textbf{ContextLens}: We prompt LLMs under the \name's pipeline to evaluate the compliance performance without any extra model tuning.


\subsection{Evaluation Metrics}

To quantify the compliance reasoning performance, we implement regular expression parsers to capture the generated predictions and regard parsing failures as incorrect.
Since the data distribution is imbalanced and every class is important, we report the accuracy and Macro-F1 as the LLMs' overall performance.
For cases with unknown contextual factors identified by our \name, we calculate the proportion of these imperfect cases among all cases, denoted as ``Ratio.''
To quantify the reliability of listed contextual factors, we conduct human evaluations and report the Cohen's kappa coefficients between human evaluators and LLMs.
In addition, we also report Fleiss' kappa for inter-LLM agreement among the 3 LLMs.

\section{Experimental Results}

\begin{table*}[t]
\centering
\small
\begin{tabular}{@{}l l|cc|c c|c c@{}}

\toprule
        \multirow{2}{*}{Evaluation Types} & 
        \multirow{2}{*}{Evaluated Models} 
        & \multicolumn{2}{c|}{\textbf{EU AI Act}} & \multicolumn{2}{c|}{\textbf{GDPR}} & \multicolumn{2}{c}{\textbf{Average}}\\

   {} & {}            &  Acc & Macro-F1 &  Acc & Macro-F1 & Acc & Macro-F1   \\
   \midrule
    \multirow{4}{*}{Direct} &
  Qwen2.5-7B-Instruct    & 46.33 & 41.89 & 88.37 & 85.36 &  67.35 & 63.63 \\
& Llama-3.1-8B-Instruct &58.67 &59.36 &86.88 &84.06 &72.78 & 71.71 \\
& OpenThinker-7B    &70.50 & 69.72 & 87.26 & 81.82 &78.88& 75.77\\

& GPT-4o-mini & 71.16 & 70.92 & \textbf{92.19} & 89.29 &  81.68& 80.11\\
& GPT-4o &   75.33 & 73.39 & \textbf{92.19} & \textbf{90.02}  & 83.76 & 81.71 \\

& GPT-5 &   74.83 & 72.07 & 87.57 & 84.33  & 81.20 & 78.20 \\

\midrule
\multirow{3}{*}{Long CoT} & DeepSeek-R1 (671B)  & 84.00 & 83.30 & 90.44 & 88.17 & 87.22 & 85.74\\
& o3-mini & 84.83 & 84.36 & 88.69 & 86.82 & 86.76 & 85.59 \\
& Gemini-2.5-Flash & 80.00 & 79.48 &  92.03 & 89.19 & 86.02 & 84.34 \\

\midrule
\multirow{3}{*}{RAG} & GPT-4o-mini  & 74.33 & 73.62 & 91.24 & 90.05 & 82.79 & 81.84\\
& o3-mini & 77.75 & 75.14 & 90.92 & 87.46 & 84.33 & 81.30\\
& GPT-5 & 80.33 & 79.80 & 91.24 & 88.37 & 85.79 & 84.09\\

\midrule
\multirow{2}{*}{Fine-tuned} & ContextReasoner-7B-SFT     &   84.33 & 83.73 &91.71& 88.36& 88.02&86.05 \\

& ContextReasoner-7B-PPO     &  84.33 & 83.65 & \textbf{92.19} & 89.33 & 88.25 & 86.49 \\
\midrule

\multirow{5}{*}{\name} & Qwen2.5-7B-Instruct & 54.63 & 53.88 &85.19 & 79.58& 70.61 \textsubscript{(+ 3.26)} & 66.73 \textsubscript{(+ 3.10)}\\
& Llama-3.1-8B-Instruct &86.60&86.21& 88.85& 85.13& 87.73 \textsubscript{(+14.95)} & 85.67 \textsubscript{(+13.96)}\\
& GPT-4o-mini & 88.11 & 87.53 & 91.87 & 88.25 & \textbf{89.99} \textsubscript{(+ 7.20)} &  \textbf{87.89} \textsubscript{(+ 6.05)}\\
& GPT-4o & 87.50 & 87.02 & 89.31 & 85.55& 88.40 \textsubscript{(+ 4.64)} &  86.29 \textsubscript{(+ 4.58)}\\
& DeepSeek-R1 (671B)  &   86.31 & 85.87 & 80.89&74.52 & 83.60 \textsubscript{(- 3.62)} & 80.20 \textsubscript{(- 5.54)}\\
& o3-mini & 85.09 & 84.48 & 85.35 & 81.09& 85.22 \textsubscript{(- 1.54)} & 82.79 \textsubscript{(- 2.80)}\\
& Gemini-2.5-Flash & \textbf{89.16} & 88.50 & 83.12 & 76.89& 86.14 \textsubscript{(+ 0.12)} & 82.70 \textsubscript{(- 1.64)} \\
& GPT-5 &   86.33 & \textbf{89.05} & 81.85 & 75.15  & 84.09 \textsubscript{(- 1.70)} & 82.10 \textsubscript{(- 1.99)}\\

\bottomrule
\end{tabular}
\vspace{-0.1in}
\caption{Overall compliance evaluation results reported under \%. Averaged differences between \name and the best-performing baselines are reported in parentheses.}
\vspace{-0.15in}
\label{tab:legal_compliance}

\end{table*}

\begin{table}[t]
\small
        \centering
\resizebox{\columnwidth}{!}{
        \begin{tabular}{l|cc|cc}
            \toprule
            & \multicolumn{2}{c|}{GDPR} 
            & \multicolumn{2}{c}{AI Act} \\
            \textbf{} & \textbf{Ratio} & \textbf{Avg \#} & \textbf{Ratio} & \textbf{Avg \#} \\
            \midrule
            Llama-3.1-8B-Instruct  &98.57 &4.93 & 98.33 & 5.22\\
            GPT-4o-mini  &99.04 &5.22 & 100.00 & 5.50\\
            Gemini-2.5-Flash  &100.00 &7.48  & 100.00 &  5.73 \\
            
            \bottomrule
        \end{tabular}
}
        \vspace{-0.1in}
        \caption{LLMs identified cases with imperfect context. ``Ratio'' measures the \% of imperfect cases. ``Avg \#'' denotes the average unknown factors per case.}
        \label{tab:imperfect_stats}
        \vspace{-0.15in}
\end{table}


\subsection{Evaluation on the Overall Compliance}

\begin{table*}[t]
\small
\centering
\begin{tabular}{l |ccc |ccc |ccc}
\toprule
 
 & \multicolumn{3}{c|}{Article 5: Principles} 
 & \multicolumn{3}{c|}{Article 6: Lawfulness} 
 & \multicolumn{3}{c}{Article 40,42: Obligations} \\
 &    Correct    &     Wrong &$\Delta$      &     Correct      & Wrong &$\Delta$ &     Correct      &    Wrong & $\Delta$    \\ 
 \midrule
 Llama-3.1-8B-Instruct &     80.82 & 95.71 &+14.89 & 75.44 & 74.28 &-1.16 & 91.03 & 95.71 &+4.68      \\
 GPT-4o-mini&     90.46 & 100.00 &+9.54 & 39.86 & 68.62 &+28.76 & 90.12 & 100.00 &+9.88   \\
 Gemini-2.5-Flash&     99.23& 100.00 &+0.77 & 58.23& 63.20 &+4.97 & 95.21& 97.16  &+1.95 \\
 \bottomrule
\end{tabular}
\vspace{-0.1in}
\caption{Proportion of imperfect cases with unknown contextual factors on specific GDPR articles.
``Correct'' denotes the imperfect ratio among all correct predictions, while ``Wrong'' refers to the ratio of wrong predictions.
$\Delta$ = ``Wrong'' - ``Correct''.}
\vspace{-0.15in}
\label{tab:GDPR_imperfect}
\end{table*}

Following the experimental setups, we compare our \name with existing baselines for the overall compliance evaluation.
As presented in Table~\ref{tab:legal_compliance}, the numerical results suggest the following findings:

1) \textit{Our \name can generally improve LLMs' overall compliance to achieve the new state-of-the-art performance.}
Notably, when prompted with \name, the averaged compliance performance of GPT-4o and GPT-4o-mini even outperforms the ContextReasoner, the best-performing baselines fine-tuned via SFT and reinforcement learning.
Furthermore, \name demonstrates remarkable efficacy on smaller LLMs. 
Compared to the Direct method, our \name yields an averaged accuracy improvement exceeding 8\% on GPT-4o-mini, outperforming its larger counterpart, GPT-4o.
In addition, Llama-3.1-8B-Instruct under \name gains more than 14\% averaged accuracy and Marco-F1 scores.

2) \textit{\name significantly enhances  LLMs' compliance performance on the AI Act subset.}
When comparing compliance performance between the EU AI Act and GDPR sub-domains, our \name achieves substantial improvements in the EU AI Act subset.
For instance, in GPT-4o-mini, the EU AI Act accuracy increases dramatically from 71.16\% in Direct to 88.11\% and Gemini-2.5-Flash obtains 9\% accuracy improvement.
In contrast, for the GDPR subset, \name fails to yield consistent improvements and, in some cases, even leads to performance degradation, particularly with long CoT LRMs.

3) \textit{Both \name and RAG are not always effective on LRMs.}
For models in the Long CoT category, such as DeepSeek-R1, o3-mini, and Gemini-2.5-Flash, RAG and \name methods even result in reduced accuracies and Macro-F1 on the GDPR subset. 
These LRMs typically organize their detailed planning into the thinking procedures, and enforcing specified rules of \name may not always be beneficial.
While Long CoT LRMs naturally have comparable compliance performance, these LRMs tend to incur significantly higher costs, such as increased token generation, slower inference speeds, higher token pricing, and a greater likelihood of parsing failures.
Therefore, our \name achieves a more efficient trade-off for directly instruction-tuned LLMs.


\subsection{Analysis on the Imperfect Context}
In addition to compliance assessment, compared with all baseline methods, our \name can identify the missing and ambiguous contextual factors.
We select GPT-4o-mini, Llama-3.1-8B-Instruct, and Gemini-2.5-Flash as representative LLMs to systematically analyze the ambiguous factors within real-life contexts in the GDPR subset and synthetic cases in the EU AI Act subset.
For the GDPR, our \name covers 32 contextual factors that may be classified as ``not sure.''
For the EU AI Act, 8 out of 10 questions include the ``None of the above'' option.
The analysis of the imperfect context reveals the following findings:

4) \textit{Almost all existing privacy and safety contexts are inherently imperfect.}
Table~\ref{tab:imperfect_stats} calculates the proportion of cases with unknown contextual factors and the averaged number of factors per case for the 3 evaluated LLMs.
The results show that nearly 100\% of cases involve more than 5 unknown factors that cannot be inferred from their current context.
Given the existence of such unknown factors, prior contextual judgments may not be comprehensive and fully reliable.
For detailed analyses on these imperfect factors, we further conduct both human-LLM agreements and cross-LLM agreements. The results are shown in Appendix~\ref{sec: Agreement}.

5) \textit{Imperfect context leads to undermined compliance assessment.}
As presented in Table~\ref{tab:GDPR_imperfect}, we investigate the impact of unknown factors on the correctness of article-level evaluations.
The results indicate that LLM judges perform worse when the context is imperfect.
Using GPT-4o-mini as an example, in the identification of contextual factors under GDPR Article 6 (Lawfulness of Processing), incorrect compliance judgments are 28.76\% more likely to involve imperfect contextual factors compared to correct judgments.

\subsection{Case Studies and Failure Analysis}
In addition to quantitative experiments, we further perform case studies to provide a qualitative evaluation.
Our \name incorporates fine-grained rules to handle the imperfect context.
However, as shown in Table~\ref{tab:legal_compliance}, Gemini-2.5-Flash results in a 9\% accuracy drop with our \name on the GDPR subset.
We select one incorrectly classified sample to investigate the cause of the failure.

\begin{table}[h]
\small
\centering
\begin{tabular}{|p{0.94\linewidth}|}
\hline
\\
\textbf{Case Context:} Microsoft updated its Privacy Policy and communicated the changes via email and on its website. All users were informed, and the updates did not reduce user rights but clarified existing practices.\\
\textbf{Ground Truth:} Permitted\\
\\
\hline
\\
\textbf{Gemini-2.5-Flash’s Structured Output}:\\
\{ \\ 
\hspace*{1em} ``Article 6(1)(a)": ``no" (Prohibited), \\
\hspace*{1em} ``Article 6(1)(b)": ``not sure" (Imperfect Context), \\
\hspace*{1em} ``Article 6(1)(c)": ``no" (Prohibited), \\
\hspace*{1em} \ldots \\
\}

\textbf{Rule-based Assessment:} \textcolor{red}{Prohibited} (Wrong Prediction)\\
\\
\hline
\\
\textbf{Gemini-2.5-Flash's Responses:}\\
\\
-- **Article 6(1)(a): Consent:** The context mentions \textcolor{cyan}{\ul{users were ``informed'' but doesn't state they explicitly ``consented'' to the processing of their data based on the updated policy}}. Simply informing users of a policy update is not the same as obtaining explicit, informed, and unambiguous consent for data processing activities.\\
-- **Article 6(1)(b): Contract:** If the privacy policy is part of the terms of service that constitute a contract with the user, and the processing of data is ``necessary for the performance of a contract,'' this could be a basis. However, \textcolor{cyan}{\ul{the context doesn't provide enough information to confirm that the personal data processing is *necessary* for the contract's performance or what the contract entails}}.\\
-- **Article 6(1)(c): Legal obligation:** The context does not indicate that the update to the privacy policy or the underlying data processing is necessary for compliance with a specific legal obligation imposed on Microsoft.\\
-- \ldots \\
\\
\hline
\end{tabular}
\vspace{-0.1in}
\caption{A case study examining a real scenario permitted under the GDPR.}
\vspace{-0.15in}
\label{tab:case_example}
\end{table}

As shown in Table~\ref{tab:case_example},
the original case context is labeled as permitted.
However, under our \name, Gemini-2.5-Flash determines that there is no lawful basis and general principle to permit the case and the case is considered as prohibited according to our rules.
Despite this incorrect prediction, our \name still enables Gemini-2.5-Flash to identify the ambiguity and incompleteness inside the case context with reasonable justifications.
The GDPR states that data processing shall be lawful only if and to the extent that at least one of Article 6(1)'s sub-items applies to the context. 
Specifically, as shown in the response, the model states that the explicit ``consent'' is not given, so that Article 6(1)(a) may not hold.
Regarding the contract necessity under Article 6(1)(b), the model expresses uncertainty due to the lack of prior contract-related information in the context.
Consequently, Article 6(1)(b) is identified as ``not sure.''
After enumerating all the sub-items for lawfulness (which is part of the \textit{applicability scope} chunk) and general principles, the model cannot find any provision to permit the case.
Therefore, \name concludes that the context violates Article 6(1)(a).

Given these fair explanations and manual inspections, it is plausible that the case context may indeed be prohibited under GDPR.
LLMs' justifications validate that existing models are aware of the privacy and safety context.
However, an imperfect context with unknown contextual factors hinders LLM judges from performing comprehensive assessments.




\begin{table}[ht]
    \centering
    \small
    \resizebox{\columnwidth}{!}{
    \begin{tabular}{lllcc}
        \toprule
        \textbf{Method} & \textbf{Dataset} & \textbf{Model} & \textbf{Input \#} & \textbf{Response \#} \\
        \midrule
        \multirow{4}{*}{\textbf{Direct}} & AI Act & GPT-4o & 238.31 & 8.36 \\
        & AI Act & GPT-4o-mini & 238.31 & 92.92 \\
        & GDPR & GPT-4o & 149.17 & 7.45 \\
        & GDPR & GPT-4o-mini & 149.17 & 68.62 \\
        \midrule
        \multirow{4}{*}{\textbf{ContextLens}} & AI Act & GPT-4o & 522.29 & 195.30 \\
        & AI Act & GPT-4o-mini & 522.29 & 194.56 \\
        & GDPR & GPT-4o & 3002.98 & 455.25 \\
        & GDPR & GPT-4o-mini & 3002.98 & 409.98 \\
        \bottomrule
    \end{tabular}
    }
    \vspace{-0.1in}
    \caption{Token cost analysis.}  
    \label{tab:cost}
    \vspace{-0.25in}
\end{table}

\section{Cost Analysis}

As mentioned in our methodology, \name substantially increases token usage per sample compared to direct prompting.
Therefore, we compare average token consumption for both direct prompting and our method for evaluations involving gpt-4o and gpt-4o-mini and report the token costs in Table~\ref{tab:cost}.

With more token budgets, our results show that \name on a "cheaper" model (GPT-4o-mini) can outperform direct prompting on a "stronger" model (GPT-4o). 
Though the input tokens increase significantly, the total financial cost of running \name on a mini model is often still lower than running a direct prompt on a stronger model.
Regarding latency, most of the increase is in input tokens. Since LLM providers process input tokens in parallel (prefill stage) and delay primarily on output generation, the latency of \name is primarily in the longer response lengths. Still, responding to no more than 500 tokens is manageable, as current agentic tasks typically require more than 1,000 output tokens with tool-calling.
\section{Conclusion}

In this paper, inspired by the challenges of imperfect context, we propose \name, a semi-rule-based framework designed to leverage both LLMs' powerful context understanding ability and legal rules' hierarchical nature to assess compliance for both privacy and safety.
In addition to deriving compliance outcomes, these pre-defined rules allow our \name to identify the ambiguity and incompleteness in the given context, facilitating more comprehensive judgments.
Our experimental results indicate that our \name achieves state-of-the-art compliance judgment without any extra fine-tuning. 
Moreover, our analysis suggests that nearly all cases of existing compliance benchmarks are imperfect with unknown contextual factors.
For future work, we call for the development of more robust evaluation datasets that incorporate real-world contextual uncertainties to better reflect the complexities of privacy and safety context and align with people's actual concerns.


\section*{Limitations}

Though our \name is always effective on the instruction-tuned LLMs, both \name and naive RAG implementations may not be helpful for Large Reasoning Models with long CoT reasoning traces.
Our experimental results also suggest that LRMs with Long CoT suffer from compliance performance degradation on the GDPR subset.

In addition, though \name is an adaptive framework without extra LLM training, \name still requires legal experts in the loop to supervise the legal document chunking process done by LLMs and improve the chunking quality.
In other words, \name cannot yet be constructed in a fully automated manner for newly emerging regulations.

\section*{Ethical Considerations}
All authors of this paper affirm their adherence to the ACM Code of Ethics and the ACL Code of Conduct. This work is primarily aimed at identifying the missing and ambiguous contextual factors to further enhance the reliability and compliance evaluation of LLM judges on data privacy and AI safety grounded on the legal norms. 
Our \name would benefit current privacy and safety research to re-examine the ambiguity within existing benchmarks and methodologies. 
Our source code will be made publicly available.

\section*{Acknowledgements}

The authors of this paper were supported by the National Key Research and Development Program of China (2025YFE0200500), the ITSP Platform Research Project (ITS/189/23FP) from ITC of Hong Kong, SAR, China, and the AoE (AoE/E-601/24-N), the RIF (R6021-20) and the GRF (16205322) from RGC of Hong Kong, SAR, China. 
The work described in this paper was conducted in full or in part by Dr. Haoran Li, JC STEM Early Career Research Fellow, supported by The Hong Kong Jockey Club Charities Trust. We also thank the support from Huawei. 
\clearpage


\bibliography{custom}

\clearpage

\appendix

\clearpage

\section{Judicial Reasoning on Data Privacy}

As stated earlier in Section~\ref{sec: motivation}, judicial reasoning in legal cases relies on a hierarchy of legal maxims, including lex superior derogat legi inferiori (higher law prevails over lower), lex specialis derogat legi generali (the specific prevails over the general), and lex posterior derogat legi priori (the later prevails over the earlier). 

Applying the lex specialis principle to the GDPR framework, a critical debate centers on the hierarchy between the specific lawfulness criteria in Article 6 and the provisions for children in Article 8 and special categories of data in Article 9. 
While some literature advocates for a cumulative application model where both specific and specific grounds must be satisfied concurrently \citep{Comandè_Schneider_2022, Florea-Withdrawal-2023}, our \name treats Article 8 and Article 9 as special exceptions that override specific rules. 
Following Kelsen’s theory of normative hierarchy \citep{Kelsen1991GeneralTO}, a norm conflict exists when a permissive rule like Article 6 interacts with a prohibitive-exception model like Article 9, requiring a test of violation to determine priority \citep{Vranes-Definition-2006}. Drawing from regulatory interpretations \citep{Georgieva-Article9-2020, EDPB2019Opinion5}, we identify three requirements for this special status: the specificity of the material scope \citep{Dove-GDPR-2018}, the intensification of normative thresholds such as explicit consent \citep{Dove-What-2021}, and the displacement of broad general bases like legitimate interest to prevent regulatory evasion \citep{Korff2024LegalBasesAI, MolnarGabor2018FairBalance}. By encoding these hierarchical priorities, \name ensures that high-risk processing contexts are governed by their respective special conditions.

\section{Experimental Details}

\paragraph{Generation Details} 

For open-source models, we follow the models' recommended configurations in their model cards.
For closed-source models, we use their official APIs to obtain the responses.
To ensure reproducibility, we use greedy search decoding and set the random\_seed = 42 with temperature = 0. 
For each generation among all models, we set the max\_new\_token = 1024 with max\_retry = 3.

\paragraph{Computational Resources} During our experiment, we use 2 NVIDIA H800 to run our codes for open-source models, and it takes 4 weeks of GPU hours to complete all experiments.
In terms of API cost, our overall cost for calling APIs is approximately \$300 USD.

\paragraph{Licenses}

Our evaluated testing data are under the CC BY-NC-SA 4.0 license and the U.S. copyright laws, and we can use them for non-commercial and research purposes.
In terms of used models, we have agreed with all their model-specific licenses for research purposes.

\begin{table}[t]
\small
        \centering
\resizebox{\columnwidth}{!}{
        \begin{tabular}{l|cc|cc}
            \toprule
            & \multicolumn{2}{c|}{GDPR} 
            & \multicolumn{2}{c}{AI Act} \\
            \textbf{} & \textbf{Cohen} & \textbf{Fleiss} & \textbf{Cohen} & \textbf{Fleiss} \\
            \midrule
            Llama-3.1-8B-Instruct  &100.00  & &95.87 \\
            GPT-4o-mini &100.00 & 21.25 &96.63 & 79.93\\
            Gemini-2.5-Flash  &100.00  & &91.02 \\
            
            \bottomrule
        \end{tabular}
        }
        \vspace{-0.1in}
        \caption{Evaluation on human-LLM and inter-LLM agreement. ``Cohen'' measures Cohen's kappa coefficient between human evaluators and LLMs. ``Fleiss'' denotes the Fleiss' kappa for inter-LLM agreement.
        All coefficients are reported with a fixed scaling factor of 100.}
        \label{tab:human}
        \vspace{-0.15in}
\end{table}

\section{More on Human Evaluation}

\subsection{Agreement of Identified Imperfect Contexts}
\label{sec: Agreement}

To validate the reliability of contextual factors identified by our \name, we further conduct expert inspections and cross-LLM comparisons. 
Two graduate students with a legal background are invited to assess the plausibility of the identified unknown contextual factors.
Within each domain and LLM, 30 cases with imperfect context are randomly chosen.
To quantify the agreement, we calculate Cohen's kappa coefficients between human experts and LLMs.
For cross-LLM consistency, we report Fleiss' kappa among the three evaluated LLMs.
In the GDPR domain, we determine the agreement at the sub-item level and report the kappa coefficients at the article level for better readability.
For the EU AI Act, the coefficients are reported at the question level.
Table~\ref{tab:human} presents the agreement coefficients for both human–LLM and inter-LLM comparisons. 
Its results support the following key findings:


\textit{LLMs can effectively identify missing contextual factors for legal compliance.}
In Table~\ref{tab:human}, the Cohen’s kappa coefficients between LLMs and human annotators exceed 0.9 for both the GDPR and the AI Act domains.
These high agreement scores indicate that LLMs are capable of analyzing the missing contextual factors and perform comparably to human annotators.

\textit{Different LLMs identify diverse contextual factors.}
As shown in Table~\ref{tab:human}, in the GDPR domain, all 3 LLMs achieve the Fleiss’s kappa of 0.21, which indicates fair agreement but not complete consensus.
Given that most LLMs can correctly identify the missing factors with fair explanations based on varied perspectives, ensembling their outputs may offer a comprehensive analysis of the given context.

\section{Legal Document Chunking Details}
\label{app:stat}

In this section, we systematically explain how we chunk the GDPR to assign \textit{applicability scope}, \textit{special conditions}, \textit{common provisions} and \textit{general principles} at the item level.
Without loss of generality, we only list the regulation content at the article level for better readability.

For the \textit{applicability scope}, we include all the sub-items inside Articles 2 and 3 to cover both the material scope and territorial scope:

\begin{tcolorbox}[colback = cBlue_1!10, colframe = cBlue_7,  coltitle=white,fonttitle=\bfseries\small, center title,fontupper=\small,fontlower=\small]

\textbf{GDPR Articles for Applicability Scope:} \\
- Article 02: Material scope \\
- Article 03: Territorial scope 
\end{tcolorbox}

For \textit{special conditions}, we manually review the main body of the GDPR to list articles that cover provisions regarding special situations and exceptions of higher priority.
Then, one legal expert is invited to validate the identified articles.
The selected special conditions are listed below:

\begin{tcolorbox}[colback = cBlue_1!10, colframe = cBlue_7,  coltitle=white,fonttitle=\bfseries\small, center title,fontupper=\small,fontlower=\small]

\textbf{GDPR Articles for Special Conditions:} \\
- Article 08: Conditions applicable to child's consent \\
- Article 09: Processing of special categories of personal data about racial or ethnic origin, political opinions, religious or philosophical beliefs, or trade union membership, genetic data, biometric data, health data, sex life or sexual orientation. \\
- Article 10: Processing of personal data relating to criminal convictions and offences. \\
- Article 11: Processing of personal data do not or do no longer require the identification of a data subject. \\
- Article 44: Transfers of personal data to third countries or international organisations. \\
- Article 86: Processing and public access to official documents. \\
- Article 87: Processing of the national identification number. \\
- Article 88: Processing in the context of employment. \\
- Article 89: Safeguards and derogations relating to processing for archiving purposes in the public interest, scientific or historical research purposes or statistical purposes 
\end{tcolorbox}

For \textit{common provisions}, we consider articles under the topics of lawfulness, processor obligations, and rights of the data subject as common provisions.
The detailed structure is shown below:

\begin{tcolorbox}[colback = cBlue_1!10, colframe = cBlue_7,  coltitle=white,fonttitle=\bfseries\small, center title,fontupper=\small,fontlower=\small]

\textbf{GDPR Articles for Common Provisions:} \\
**GDPR Articles for Lawful Basis:** \\
- Article 06: Lawfulness of processing \\
**GDPR Articles for Processor/Controller Obligations:**\\
- Article 40: Codes of conduct \\
- Article 42: Certification \\
**GDPR Articles for Subject Rights:** \\
- Article 13: Information to be provided where personal data are collected from the data subject. \\
- Article 14: Information to be provided where personal data have not been obtained from the data subject. \\
- Article 15: Right of access by the data subject. \\
- Article 16: Right to rectification. \\
- Article 17: Right to erasure ('right to be forgotten'). \\
- Article 18: Right to restriction of processing. \\
- Article 20: Right to data portability. \\
- Article 21: Right to object. \\
- Article 22: Automated individual decision-making, including profiling.
\end{tcolorbox}

For \textit{general principles}, we cover all the items and sub-items under article 5.

\begin{tcolorbox}[colback = cBlue_1!10, colframe = cBlue_7,  coltitle=white,fonttitle=\bfseries\small, center title,fontupper=\small,fontlower=\small]

\textbf{GDPR Articles for General Principles:} \\
- Article 05: Principles relating to processing of personal data 
\end{tcolorbox}

\begin{table}[t]

\small
\centering
\begin{tabular}{p{0.43\textwidth}}
\toprule

\textbf{Multiple Choice Template Used for Direct Prompting}\\
    Determine the relationship between the provided event and the \textcolor{contentcolor}{\texttt{<domain>}} regulations. You have three options for your conclusion:\\
(A) The event is prohibited by \textcolor{contentcolor}{\texttt{<domain>}}.\\
(B) The event is permitted by \textcolor{contentcolor}{\texttt{<domain>}}.\\
(C) The event is not related to \textcolor{contentcolor}{\texttt{<domain>}}.\\

Event:
\textcolor{contentcolor}{\texttt{<event>}}

Output Format:

Choice: [A. Prohibited \texttt{|} B. Permitted \texttt{|} C. Not related ]
\\
\bottomrule
\end{tabular}
\caption{Evaluated multiple-choice prompt template for direct prompting. Light blue texts inside each ``\textcolor{contentcolor}{\texttt{<>}}'' block denote a string variable.}
\label{app-tab: prompt_template_direct}
\end{table}

\begin{table*}[t!]

\small
\centering
\begin{tabular}{p{2\columnwidth}}
\toprule

\textbf{ContextLens’ Context Analysis Prompt Template for the EU AI Act}\\
As an expert context analyzer, your task is to analyze the given case and identify if there is any AI system involved. If there is AI system involved, please identify the name, type, usage and actions of the AI system. \\

**Definition of AI system**: \\
An AI system is defined as: A machine-based system designed to operate with varying levels of autonomy and that may exhibit adaptiveness after deployment and that, for explicit or implicit objectives, infers, from the input it receives, how to generate outputs such as predictions, content, recommendations, or decisions that can influence physical or virtual environments. \\

Please follow these steps: \\
1. Analyze the case and identify if there is any AI system involved. If there is an AI system involved, please identify the name of the AI system, the type of the AI system, and the usage of the AI system. \\
2. If there is an AI system involved, please identify atomic actions that are performed by the AI system and the target of the action. If there is more than one action, please identify all of them. \\
3. For each action, please identify the target of the action and the purpose of the action.\\

**Case**: 
\textcolor{contentcolor}{\texttt{<event>}} \\

**Output format**:
Output format should be in JSON format: \\
\{ \\
\hspace*{1em}        ``AI\_system\_involved": True/False, \\
\hspace*{1em}        ``AI\_system\_name": ``name of the AI system", \\
\hspace*{1em}           ...\\
\} \\

\midrule

\textbf{ContextLens' Subsequent Prompt Template for the EU AI Act if AI System is Involved}\\
Your task is to play the role of the AI system inside the case. Please answer the following questions based on the case content and the case context. For each question, there may be more than 1 applicable option, you should provide a list of options. \\

**Case content**: \textcolor{contentcolor}{\texttt{<event>}}\\
**Case context**: \textcolor{contentcolor}{\texttt{<Analyzed Context>}}\\
**List of questions**:\\
**Question 1**: Which kind of entity is your organisation? \\
\qquad**Options**:\\
\qquad1. Provider\\
\qquad2. Deployer\\
\qquad3. Distributor\\
\qquad4. Importer\\
\qquad5. Product manufacturer\\
\qquad6. Authorised representative\\
**background**: Definitions:\\
Provider: a natural or legal person, public authority, agency or other body that develops an AI system or a general purpose AI model (or that has an AI system or a general purpose AI model developed) and places them on the market or puts the system into service under its own name or trademark, whether for payment or free of charge;\\
Deployer: any natural or legal person, public authority, agency or other body using an AI system under its authority except where the AI system is used in the course of a personal non-professional activity;\\
...\\
Product manufacturer: places on the market or puts into service an AI system together with their product and under their own name or trademark;\\
**Question 2**: 
Has a downstream deployer, distributer, or importer made any of the following modifications to your system?\\
\qquad**Options**:\\
\qquad1. Putting a different name/trademark on the system\\
\qquad2. Modifying the intended purpose of a system already in operation\\
\qquad3. Performing a substantial modification (see Article 3 point 23) to the system\\
\qquad4. None of the above\\

...\\

*Question 10**:Does your AI system (or the product for which your AI system is a 'safety component') fall within any of the following high-risk categories?\\
\qquad**Options**:\\
\qquad1. Civil aviation security\\
\qquad...\\
\qquad8. None of the above\\

**Output format**: Output format should be in JSON format and your answer should contain only the numerical index of the option without any text in the options. You should select at least one option for each question.\\
\{ \\
\hspace*{1em} ``question\_1": [option\_1, option\_2, ...], \\
\hspace*{1em} ``question\_2": [option\_1, option\_2, ...], \\
\hspace*{1em} ...\\
\}\\

\bottomrule
\end{tabular}
\caption{Evaluated prompt templates for ContextLens. Texts inside each ``\textcolor{contentcolor}{\texttt{<>}}'' block denote a string variable.}
\label{app-tab: prompt_template_ai_act}
\end{table*}

\section{Prompt Templates}

In this section, we list our detailed prompt templates used for direct prompting and \name pipeline as well as their token costs.
As prompts used in our \name are lengthy, we only show our prompts on the EU AI Act subset for simplicity.

\paragraph{Multiple-choice Prompting Template}

Our baseline methods leverage the direct prompting method to answer the multiple-choice question to assess the compliance performance, which is shown in
Table~\ref{app-tab: prompt_template_direct}.
The direct prompting template asks evaluated LLMs the compliance question with the given case context.
Then, we implement the parser to parse the corresponding options to get the models' predictions.
The average token cost is approximately 350 tokens per sample.

\paragraph{Templates Used by ContextLens}

Instead of directly asking LLMs' compliance assessment, our \name prompts LLMs with a list of questions to align the cases' context with the regulatory statutes.
As shown in Table~\ref{app-tab: prompt_template_ai_act}, for the EU AI Act domain, our \name first identifies if any AI system is involved to determine applicability.
Then, if applicable, our \name enumerates a list of contextual questions to fit into the AI Act's criteria.
After assigning the options for the questions, we additionally implement a rule-based method to automatically derive the compliance outcome without LLMs.
The average length of \name's prompts is more than 2,000 tokens for each given case.


\section{How LRMs Fail on \name}


In this section, we perform more qualitative analysis to discuss why large reasoning models fail on our \name pipeline.

First, large reasoning models prefer to consider more contextual factors as ambiguous.
For example, as shown in Table~\ref{tab:imperfect_stats}, Gemini-2.5-Flash tends to mark more cases as imperfect with more averaged unknown factors for both the GDPR and AI Act.
In GDPR, Gemini-2.5-Flash lists more than 7 averaged unknown factors for each case. 
This number suggests that LRMs are too conservative with excessive uncertainty.

Second, LRMs heavily overthink on their reasoning traces for ambiguous contextual factors.
As shown in Table~\ref{tab:app-lrm}, for Article 6(1)(a), DeepSeek-R1 is confident that the context should be prohibited by this sub-item as the consent is not valid.
However, when it comes to Article 6(1)(b), DeepSeek-R1 becomes overly cautious and hesitant with a long thinking process filled with ``but'' and ``wait''.
This shift suggests that LRMs struggle to maintain consistency and confidence through their reasoning.
What is worse, this type of reasoning results in low-quality overthinking with repetition and confusion.
Consequently, LRMs fail to improve their compliance performance under our \name pipeline.

\subsection{Uncertainty Harms LRMs}

For the design of \name, we set the default prediction to 'prohibited' when no permissive basis can be confirmed for our aggregation logic. In legal compliance assessment, we believe a False Positive (erroneously permitting a violation) is significantly more damaging than a False Negative (erroneously flagging a violation). If a Permissive Basis cannot be confirmed with high confidence, we consider it as prohibited by default. However, a prohibited result triggered by uncertainty should require further human audit. It is not a definitive error, but a signal that the current evidence is insufficient to grant the permission.

To exclude the influence of no permissive basis, we exclude the indeterminate cases (Default \#) with no permissive and prohibitive basis. The result is shown in Table~\ref{tab:permissive basis}.
After excluding the indeterminate cases, we recalculate the updated accuracy and observe consistent improvements over the original accuracy. Moreover, LRMs tend to identify more indeterminate cases than instruction-tuned LLMs. By excluding the indeterminate cases, LRMs obtain more accuracy gain. This accuracy improvement suggests that uncertainty without concrete legal grounding actually harms the compliance assessment performance, especially for LRMs.

\begin{table}[ht]
    \centering
    \small
    \resizebox{\columnwidth}{!}{
    \begin{tabular}{lccc}
        \toprule
        \textbf{Model} & \textbf{Default \#} & \textbf{Original Acc} & \textbf{Updated Acc} \\
        \midrule
        Qwen2.5-7b-Instruct   & 22 & 85.19 & 85.64 \\
        Llama-3.1-8B-Instruct & 8  & 88.85 & 89.19 \\
        GPT-4o                & 19 & 89.31 & 90.13 \\
        DeepSeek-R1 (671B)    & 26 & 80.89 & 83.22 \\
        o3-mini               & 24 & 85.35 & 86.42 \\
        Gemini-2.5-Flash      & 31 & 83.12 & 85.09 \\
        \bottomrule
    \end{tabular}
    }
    \vspace{-0.1in}
    \caption{Model performance with and without permissive basis. Default \# refers to the indeterminate cases with no permissive and prohibitive basis.}  
    \label{tab:permissive basis}
    \vspace{-0.25in}
\end{table}

\begin{table}[h]
\centering
\small
\label{tab:evaluation-results}
\begin{tabular}{llcc}
\toprule
\textbf{RAG Types} & \textbf{Models} & \textbf{EU AI Act} & \textbf{GDPR} \\ \midrule
\multirow{3}{*}{LegalBERT} & GPT-4o-mini & 69.67 & 92.20 \\
 & GPT-4o & 78.83 & 90.76 \\
 & o3-mini & 34.33 & 69.11 \\ 
 \midrule
 \multirow{3}{*}{BM25} & GPT-4o-mini & 74.33 & 91.24 \\
 & GPT-4o & 77.75 & 90.92\\
 & o3-mini & 80.33 & 91.24 \\ 
 \bottomrule
\end{tabular}
\vspace{-0.1in}
\caption{Accuracy evaluation with different RAG implementations.}
\label{tab:rag-compare}
\vspace{-0.25in}
\end{table}

\section{Evaluation on Different RAG Setups}

To further investigate the performance of RAG, we select a legal-tuned retriever, LegalBERT~\cite{chalkidis-etal-2020-legal}, for the RAG setup.
Table~\ref{tab:rag-compare} presents the overall compliance accuracy evaluation results of the LegalBERT-based RAG system.
The results suggest that using a retriever optimized for legal documents does not consistently lead to significant improvements in downstream RAG performance. 
There may also exist a domain gap: as the EU AI Act is a relatively new regulation, employing a retriever trained on prior legal corpora may even lead to a degradation in effectiveness.

\section{Prompt Sensitivity Analysis}

As discussed in the experimental settings, we set the temperature = 0 with a fixed random seed to ensure reproducibility. 
To further enhance our statistical significance, we repeat our method 3 times with different random seeds on gpt-4o and gpt-4o-mini. We report the sample standard deviation and 95\% confidence interval for accuracy in Table~\ref{tab:sensitivity}.

\begin{table}[ht]
    \centering
    \small
    \resizebox{\columnwidth}{!}{
    \begin{tabular}{llcc}
        \toprule
        \textbf{Dataset} & \textbf{Model} & \textbf{Accuracy ($\pm$ SD)} & \textbf{95\% CI} \\
        \midrule
        AI\_ACT & gpt-4o & $87.38\% \pm 0.25\%$ & $[86.75, 88.01]$ \\
        AI\_ACT & gpt-4o-mini & $89.28\% \pm 0.25\%$ & $[88.65, 89.91]$ \\
        \addlinespace
        GDPR & gpt-4o & $86.25\% \pm 2.56\%$ & $[79.89, 92.61]$ \\
        GDPR & gpt-4o-mini & $91.88\% \pm 0.16\%$ & $[91.48, 92.27]$ \\
        \bottomrule
    \end{tabular}
    }
    \vspace{-0.1in}
    \caption{Model Performance Comparison (Values represent Accuracy $\pm$ Sample SD)}  
    \label{tab:sensitivity}
    \vspace{-0.25in}
\end{table}


\begin{table}[t]
\small
\centering
\begin{tabular}{|p{0.94\linewidth}|}
\hline
\\
\textbf{Case Context:} A doctor operates a website to provide information and services to patients and potential patients. However, the website lacks a privacy policy, which is necessary to inform users about how their personal data is collected, used, and protected. This omission means that users are not informed about the handling of their personal data when they visit the website.\\
\textbf{Ground Truth:} prohibited\\
\\
\hline
\\
\textbf{DeepSeek-R1’s Structured Output}:\\
\{ \\ 
\hspace*{1em} ``Article 6(1)(a)": ``no" (Prohibited), \\
\hspace*{1em} ``Article 6(1)(b)": ``yes" (Permitted - One of the Lawful Basis), \\
\hspace*{1em} ``Article 6(1)(c)": ``no" (Prohibited), \\
\hspace*{1em} \ldots \\
\}

\textbf{Rule-based Assessment:} \textcolor{red}{Permitted} (Wrong Prediction)\\
\\
\hline
\\
\textbf{DeepSeek-R1's Responses:}\\
\\
-- **Article 6(1)(a): Consent:** If the website is processing data without a privacy policy, users can't give informed consent because they aren't told how their data is used. So consent isn't valid here. So this would be "no".\\
-- **Article 6(1)(b): Contract:**If the processing is necessary for a contract. The website provides services, maybe users sign up for something. \textcolor{cyan}{\ul{But without a privacy policy}}, it's unclear if the data processing is necessary for the contract. \textcolor{cyan}{\ul{But the main issue is that even if they rely on (b), they still need to inform users via a privacy policy.}} 
However, the question is about the lawful basis existing, not the transparency requirement. If the processing is necessary for the service, then (b) might apply. \textcolor{cyan}{\ul{But the context doesn't specify what data is processed or for what purpose.}}\\
...\\
\textcolor{cyan}{\ul{But wait, the problem is that the doctor's website might be processing data beyond what's necessary for the contract. }}\\
...\\
Alternatively, the lack of a privacy policy doesn't affect the lawful basis. The doctor could be processing data under (b) for appointment requests, \textcolor{cyan}{\ul{but failing to inform users.}} So the lawful basis is (b), \textcolor{cyan}{\ul{but they are violating transparency.}} Therefore, the answer would be Article 6(1)(b): "yes", others "no". 
\\
\\
\hline
\end{tabular}
\vspace{-0.1in}
\caption{A case study examining a real scenario prohibited under the GDPR.}
\vspace{-0.15in}
\label{tab:app-lrm}
\end{table}

\end{document}